\documentclass[letterpaper]{article} % DO NOT CHANGE THIS
\usepackage{aaai25}  % DO NOT CHANGE THIS
\usepackage{times}  % DO NOT CHANGE THIS
\usepackage{helvet}  % DO NOT CHANGE THIS
\usepackage{courier}  % DO NOT CHANGE THIS
\usepackage[hyphens]{url}  % DO NOT CHANGE THIS
\usepackage{graphicx} % DO NOT CHANGE THIS
\usepackage{natbib}  % DO NOT CHANGE THIS AND DO NOT ADD ANY OPTIONS TO IT
\usepackage{caption} % DO NOT CHANGE THIS AND DO NOT ADD ANY OPTIONS TO IT
\usepackage{algorithm}
\usepackage{algorithmic}

\usepackage{booktabs}
\usepackage{pifont}  % For ding symbols
\usepackage[dvipsnames]{xcolor} % For color definitions
\usepackage{amsmath} 
\usepackage{multirow}

\usepackage{newfloat}
\usepackage{listings}
\DeclareCaptionStyle{ruled}{labelfont=normalfont,labelsep=colon,strut=off} % DO NOT CHANGE THIS
\floatstyle{ruled}
\newfloat{listing}{tb}{lst}{}
\floatname{listing}{Listing}
\title{InstantSticker: Realistic Decal Blending via Disentangled Object Reconstruction}
\author{
    Yi Zhang\textsuperscript{\rm 1}, 
    Xiaoyang Huang\textsuperscript{\rm 1}, 
    Yishun Dou\textsuperscript{\rm 2}, 
    Yue Shi\textsuperscript{\rm 1}, 
    Rui Shi\textsuperscript{\rm 1}, 
    Ye Chen\textsuperscript{\rm 1}, \\
    Bingbing Ni\textsuperscript{\rm 1}\thanks{Corresponding author: Bingbing Ni.}, 
    Wenjun Zhang\textsuperscript{\rm 1}
}
\affiliations{
    \textsuperscript{\rm 1}Shanghai Jiao Tong University, Shanghai 200240, China \\
    \textsuperscript{\rm 2}Huawei \\
    \{yizhangphd, nibingbing\}@sjtu.edu.cn
}

\usepackage{bibentry}
\begin{document}

\maketitle

\begin{abstract}
We present InstantSticker, a disentangled reconstruction pipeline based on Image-Based Lighting (IBL), which focuses on highly realistic decal blending, simulates stickers attached to the reconstructed surface, and allows for instant editing and real-time rendering. To achieve stereoscopic impression of the decal, we introduce \textit{shadow factor} into IBL, which can be adaptively optimized during training. This allows the shadow brightness of surfaces to be accurately decomposed rather than baked into the diffuse color, ensuring that the edited texture exhibits authentic shading. To address the issues of warping and blurriness in previous methods, we apply As-Rigid-As-Possible (ARAP) parameterization to pre-unfold a specified area of the mesh and use the local UV mapping combined with a neural texture map to enhance the ability to express high-frequency details in that area. For instant editing, we utilize the Disney BRDF model, explicitly defining material colors with 3-channel diffuse albedo. This enables instant replacement of albedo RGB values during the editing process, avoiding the prolonged optimization required in previous approaches. In our experiment, we introduce the Ratio Variance Warping (RVW) metric to evaluate the local geometric warping of the decal area. Extensive experimental results demonstrate that our method surpasses previous decal blending methods in terms of editing quality, editing speed and rendering speed, achieving the state-of-the-art.
\end{abstract}


\begin{figure}[!t]
\centering
\includegraphics[width=0.98\columnwidth]{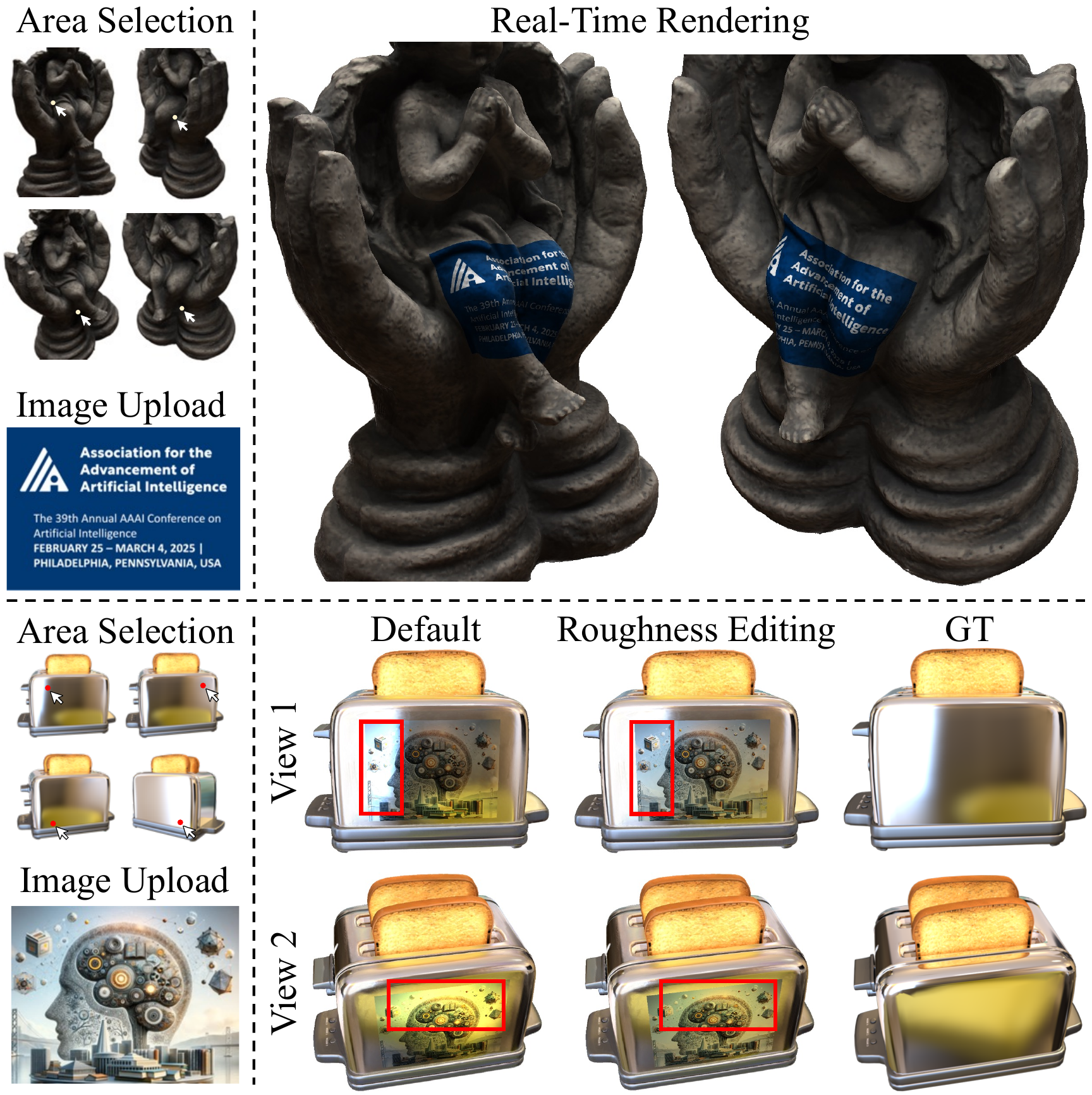}
% Reduce the figure size so that it is slightly narrower than the column. Don't use precise values for figure width. This setup will avoid overfull boxes.
\caption{
We select a target area in the user interface to position the decal by adjusting the viewpoint and selecting four anchors to define a quadrilateral. 
Then, we upload an image, which will blend realistically within this quadrilateral. 
}
\label{fig:teaser}
\end{figure}

\section{Introduction}
Decal blending for multi-view reconstructed objects allows users to upload an image and apply it to a specified area on the 3D object.
This is a flexible and controllable editing mode that creates highly realistic visual effects, making it valuable for applications in AR/VR, advertising, and digital content creation.
However, highly coupled 3D representations such as NeRF and its variants \cite{mildenhall2020nerf, tang2022compressible}, and 3D Gaussian \cite{kerbl20233d} often perform poorly in this task.
To achieve realistic decal appearance, it should conform to the surface’s bumps and depressions while accurately capturing environmental lighting and shadows, which necessitates disentangled reconstruction.

Since simultaneously disentangling all components, including geometry, often results in convergence to local minima, most methods focus on disentangling appearance components based on pre-reconstructed geometry. 
In DE-NeRF \cite{wu2023nerf}, a signed distance function (SDF) is first trained on multi-view images, followed by mesh reconstruction using the marching cubes algorithm \cite{lorensen1987marching}. 
Each vertex is assigned learnable geometry and appearance features, with lighting approximated by a learnable environment map and an implicit lighting network.
During rendering, multiple points are sampled along each camera ray, with each point’s feature derived from the weighted sum of its $k$ nearest vertices’ features, leading to significant computational overhead.
Volume rendering is then performed along each ray, using multi-view images as supervision to optimize vertex features and lighting.
During editing, a manually edited single-view rendered image is used as the optimization target. 
Raycasting from the camera to the reconstructed mesh locates the corresponding vertices, and the editing effect is achieved by iteratively optimizing the features of these vertices. 
However, this process introduces significant latency in the editing workflow and causes decal warping on curved surfaces.
Additionally, the decal often appears blurred due to pixel colors being influenced by multiple neighboring vertices, limiting the texture frequency.

In this paper, we develop a novel disentangled reconstruction pipeline based on IBL \cite{debevec1998rendering}, enabling real-time rendering and supporting instant, highly realistic decal blending with integrated environmental lighting and shadows.
Unlike DE-NeRF, we employ advanced surface reconstruction methods that provide accurate geometry, eliminating the need for calculations beyond the first-hit intersections. 
This enables us to adopt mesh rasterization for real-time rendering, which in turn allows us to leverage barycentric interpolation for estimating fragment properties from vertex features, further enhancing rendering efficiency.
For material representation, we use the Disney BRDF model \cite{burley2012physically}, which integrates Lambertian \cite{pharr2016physically} for diffuse reflection and microfacet BRDF \cite{cook1982reflectance} for specular reflection.
The corresponding texture and material features are assigned to vertices for optimization, which is more efficient than the code-decoder framework.
Additionally, the Lambertian BRDF uses a 3-channel albedo to model the diffuse component, enabling instant RGB value replacement during editing.
To address warping issues in decal blending, we enable users to select an area of the mesh for ARAP parameterization \cite{liu2008local}. 
In this area, the albedo is represented using a UV mapping combined with a high-resolution neural texture map, rather than vertex features. 
This texture map is generated by a fixed UV coordinate grid combined with a multi-layer perceptron (MLP), effectively addressing the issue of insufficient texture space learning due to gaps in the training samples.

Furthermore, we tackle the limitation of IBL's inability to model shadows. 
Specifically, IBL only accounts for direct lighting, which leads to shadows being baked into the albedo during reconstruction, resulting in edited albedo that lacks realistic shadow effects.
To address this, we introduce \textit{shadow factor} into IBL rendering equation, which can be adaptively optimized in the pre-selected area during reconstruction.
Since the diffuse albedo in the pre-selected area is generated by an MLP, which is well-known for its strong spectral bias towards producing lower frequency, smoother textures \cite{rahaman2019spectral},
the shadow factor is capable of capturing higher frequency shadows without global frequency limitations.
Moreover, to improve the reconstruction quality of specular objects, we design a \textit{local reflection network} to handle reflections within the object that are local and cannot be modeled by global environment maps.

In our experiment, we develop a Ratio Variance Warping (RVW) metric to evaluate decal warping by measuring the variance in scaling ratios between texture space Euclidean distances and mesh surface geodesic distances.
We benchmark our method against other advanced techniques on the NeRF-Synthetic \cite{mildenhall2020nerf}, ShinyBlender \cite{verbin2022ref}, and DTU \cite{jensen2014large} datasets.
Compared to the state-of-the-art, our method achieves comparable reconstruction quality with significantly less warping in decal blending on curved surfaces, more realistic shadow effects, and an 800$\times$ increase in rendering speed (80 FPS vs. 0.1 FPS).
Additionally, we implement a simple user interface (UI). 
As shown in Fig. \ref{fig:teaser}, the decal appears like a sticker on diffuse surfaces, while on metal surfaces, it blends with environment specular reflections by default. 
We can also achieve a diffuse effect on metal surfaces by manually increasing the roughness value of the blending area, making it resemble a sticker.

\begin{figure*}[!t]
\centering
\includegraphics[width=0.98\textwidth]{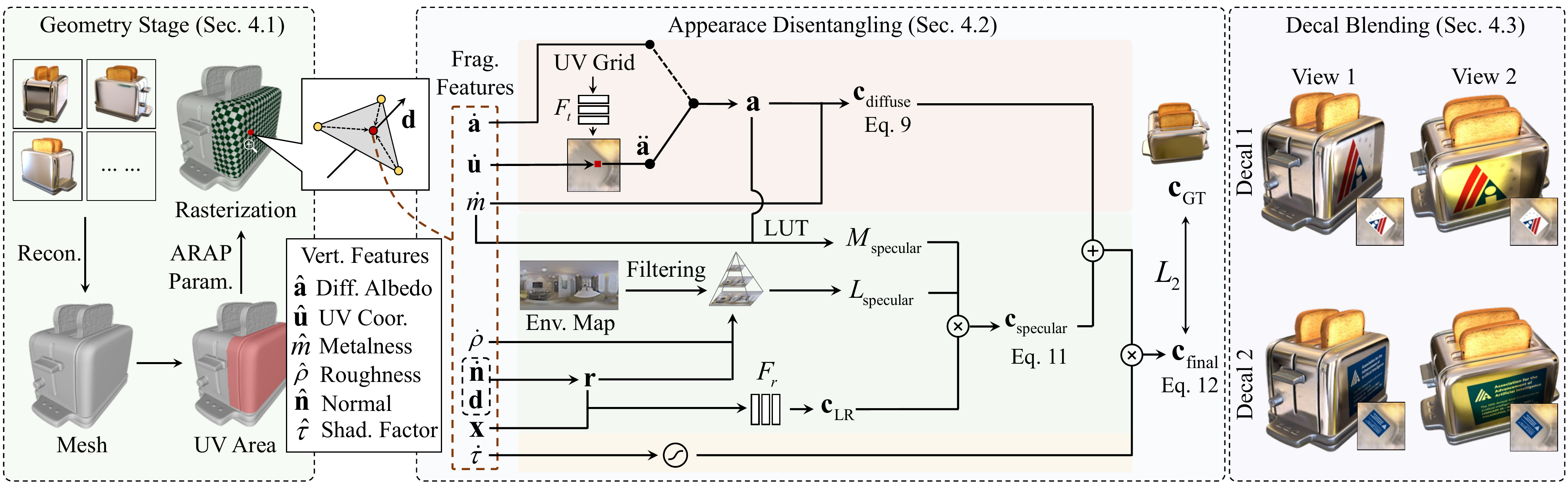}
\caption{
% During the geometry stage, we reconstruct the mesh, parameterize a UV area, and rasterize the mesh. In appearance disentangling, we assign learnable features to each vertex and use barycentric interpolation to obtain fragment features. Environment lighting is represented as a mipmap, filtered from a learnable environment map. After reconstruction, users can upload an image and blend it into any position in the UV area by overwriting its pixel colors on the albedo texture map.
During the geometry stage, we reconstruct the mesh, select and parameterize a UV area, and rasterize the mesh.
During appearance disentangling, we assign learnable features to each vertex and use barycentric interpolation to obtain fragment features. The environment lighting is represented as a mipmap, filtered from a learnable environment map. 
After reconstruction, users can upload an image and blend it into any position in UV area by overwritting its pixel colors to the albedo texture map.
Please note the distinction between different forms of the same symbol. 
For example, the symbol $\mathbf{a}$ represents the general form used in equations, $\hat{\mathbf{a}}$ denotes the learnable vertex feature, and $\dot{\mathbf{a}}$ indicates the fragment feature interpolated from $\hat{\mathbf{a}}$.
}
\label{fig:model}
\end{figure*}

\section{Related Work}
\paragraph{Mesh Parameterization.}
Mesh parameterization is a fundamental technique in computer graphics and geometric processing, often used for texture mapping, remeshing, and other applications. 
% LSCM \cite{levy2002least} aims to preserve angles, thereby ensuring the parameterization is conformal. 
% It minimizes a quadratic energy function derived from the conformal mapping equations, resulting in a linear system that can be solved efficiently.
LSCM \cite{levy2002least} approximates conformal parameterization by solving a linear system derived from conformal mapping equations.
% ARAP parameterization \cite{liu2008local} aims to minimize local distortions by preserving the local geometry as much as possible. This method iteratively solves a series of linear systems to minimize an energy function that penalizes deviations from original edge lengths.
ARAP parameterization \cite{liu2008local} minimizes local distortions by preserving local geometry as much as possible, iteratively solving linear systems to minimize an energy penalizing deviations from original edge lengths.
NeuTex \cite{xiang2021neutex} introduces an inverse mapping network and a cycle consistency loss to learn a 3D-to-2D texture mapping network for neural volumetric rendering.
Nuvo \cite{srinivasan2023nuvo} presents a UV mapping method for 3D reconstructed geometry, using a neural field to create a continuous UV map optimized for visible points, resulting in editable UV mappings that are robust to ill-behaved geometry.

\paragraph{Material and Environment Estimation.}
PhySG \cite{zhang2021physg} models geometry as an SDF network and its lighting is approximated by a
composition of several Spherical Gaussian (SG) functions.
NvDiffRec \cite{munkberg2022extracting} jointly decouples the geometry, texture, materials, and lighting from multi-view images, allowing these components to be deployed in traditional graphics engines.
NeRO \cite{liu2023nero} uses accurate sampling to recover the environment lighting and the BRDF of objects while keeping the object geometry fixed.
SOL-NeRF \cite{sun2023sol} uses a single SG lobe for sunlight approximation and models skylight with spherical harmonics, while shadows are calculated via raycasting and ambient occlusion.
GIR \cite{shi2023gir} uses 3D Gaussians to estimate the material, illumination, and geometry of an object from multi-view images.

\paragraph{Decal Blending for Reconstructed Objects.} 
NeuMesh \cite{yang2022neumesh} encodes disentangled geometry and texture on mesh vertices, enabling various editing functions, including decal blending. 
Building on NeuMesh, DE-NeRF \cite{wu2023nerf} decouples appearance into texture and lighting, enabling decals to reflect environment lighting effects. 
Seal-3D \cite{wang2023seal} achieves shadow-like decal blending by transferring luminance offsets in HSL space from the original surface color to the target surface color. 
However, these methods rely on overlaying an image on single-view renderings to achieve decal blending, which often leads to warping across different views due to the optimization being limited to a single viewpoint.

\section{Preliminary}\label{sec:pre}
% borrow from NeRO
\paragraph{Image-Based Lighting.}
IBL is a widely used real-time lighting technique in computer graphics.
% The key idea of Image-Based Lighting (IBL) is to use real-world images, typically high-dynamic-range images (HDRIs), to light a 3D scene.
The core idea of IBL is to use an HDR environment map to simulate environment lighting from an infinite distance.
% Image based lighting is shown as follows
The outgoing radiance $\mathbf{c}\left(\omega_o\right)$ in direction $\omega_o$ is calculated by
\begin{equation}\label{ibl_integral}
\mathbf{c}\left(\omega_o\right)=\int_{\Omega} L\left(\omega_i\right) f\left(\omega_i, \omega_o\right)\left(\omega_i \cdot \mathbf{n}\right) d \omega_i,
\end{equation}
which is an integral of the product of the incident radiance $L\left(\omega_i\right)$ from direction $\omega_i$ and the BRDF $f\left(\omega_i, \omega_o\right)$.
% The integration domain is the hemisphere $\Omega$ around the surface intersection normal $\mathbf{n}$.
The integration domain is the hemisphere $\Omega$ around the surface normal $\mathbf{n}$ at the ray-surface intersection.
\paragraph{Disney BRDF.}
We use the Disney BRDF model \cite{burley2012physically}, which integrates Lambertian \cite{pharr2016physically} for diffuse reflection and microfacet BRDF \cite{cook1982reflectance} for specular reflection as follows 
% Following \cite{burley2012physically}, we adopt Lambertian BRDF \cite{pharr2016physically} for diffuse term and microfacet BRDF for specular term \cite{cook1982reflectance} as follows 
\begin{equation}\label{micro_brdf}
f\left(\omega_i, \omega_o\right)=\underbrace{(1-m) \frac{\mathbf{a}}{\pi}}_{\text {diffuse}}+\underbrace{\frac{D F G}{4\left(\omega_i \cdot \mathbf{n}\right)\left(\omega_o \cdot \mathbf{n}\right)}}_{\text {specular}},
\end{equation}
where $m \in[0,1]$ and $\mathbf{a} \in[0,1]^3$ represent the metalness and diffuse albedo of the shading point, respectively.
$D, F$ and $G$ represent the GGX normal distribution function \cite{walter2007microfacet}, Fresnel term and geometry attenuation, respectively. They are all determined by the diffuse albedo $\mathbf{a}$, the metalness $m$, and the roughness $\rho \in[0,1]$, which is detailed in the appendix.
Substitute Eq. \ref{micro_brdf} into Eq. \ref{ibl_integral} to get
\begin{equation}\label{eq:3}
\mathbf{c}\left(\omega_{\boldsymbol{o}}\right)=\mathbf{c}_{\text {diffuse}}+\mathbf{c}_{\text {specular}},
\end{equation}
in which
\begin{equation}\label{eq:diff_color}
\mathbf{c}_{\text {diffuse}}
% =\int_{\Omega}(1-m) \frac{\mathbf{a}}{\pi} L\left(\omega_i\right)\left(\omega_i \cdot \mathbf{n}\right) d \omega_i
=\mathbf{a}(1-m) \int_{\Omega} L\left(\omega_i\right) \frac{\omega_i \cdot \mathbf{n}}{\pi} d \omega_i,
\end{equation}
% and
\begin{equation}\label{eq:spec_color}
\mathbf{c}_{\text {specular}}=\int_{\Omega} \frac{D F G}{4\left(\omega_i \cdot \mathbf{n}\right)\left(\omega_o \cdot \mathbf{n}\right)} L\left(\omega_i\right)\left(\omega_i \cdot \mathbf{n}\right) d \omega_i.
\end{equation}
\paragraph{Split-Sum Approximation.}
For Eq. \ref{eq:spec_color}, we use the split-sum approximation \cite{karis2013real} to simplify the integral and achieve real-time rendering via precomputation
\begin{equation}\label{eq:spec_split}
\mathbf{c}_{\text {specular}} \approx \underbrace{\int_{\Omega} L\left(\omega_i\right) D(\omega_i, \omega_0;\rho) d \omega_i}_{L_{\text {specular}}} \cdot \underbrace{\int_{\Omega} \frac{D F G}{4\left(\omega_0 \cdot \mathbf{n}\right)} d \omega_i}_{M_{\text {specular}}},
\end{equation}
where $L_{\text {specular}}$ denotes the integral of the lights on the normal distribution function $D(\omega_i, \omega_0;\rho) \in[0,1]$, which can be precomputed and stored in a mipmap.
% $\mathbf{r}$ is the reflective direction, and 
$M_{\text {specular}}$ is the integral of the specular BRDF, which can be approximated using a precomputed look-up table (LUT) as
% as shown in Fig. PUT IN APPENDIX?
% Note that a rougher surface has a larger specular lobe while a smoother surface has a smaller lobe. 
% The integral of specular BRDF can be computed by $M_{\text {specular}}=(0.04\cdot(1-m) +m \cdot \mathbf{a}) \cdot F_1+F_2$, where $F_1$ and $F_2$ are two pre-computed scalars depending on the roughness $\rho$, the view direction $\omega_o$ and the normal $\mathbf{n}$ as introduced in more details in Appendix.
% The integral of the specular BRDF can be simplified to 
\begin{equation}
M_{\text{specular}}= (0.04\cdot(1-m) +m \cdot \mathbf{a}) \cdot F_1 + F_2,
\end{equation}
where $F_1$ and $F_2$ are two precomputed scalars 
that depend on the roughness $\rho$ and the parameters $\cos\theta = \omega_i\cdot\mathbf{n}$.
% as detailed in the appendix.
The integral in Eq. \ref{eq:diff_color} can likewise be precomputed, but we have another way, which will be described in Sec. \ref{sec:AppearanceDisentangling}.

\section{Method}
% \subsection{Overview}
Our method consists of three stages, as illustrated in Fig. \ref{fig:model}. 
In Sec. \ref{sec:geo_stage}, we reconstruct the mesh and assign learnable appearance features to each vertex based on the material model discussed in preliminary, and explain how to calculate the appearance features and UVs of the shading point for each pixel.
In Sec. \ref{sec:AppearanceDisentangling}, we discuss the limitations of directly applying Eq. \ref{eq:3} along with Eqs. \ref{eq:diff_color} and \ref{eq:spec_split}, and introduce our improved disentangling pipeline. 
Finally, in Sec. \ref{sec:decal_blending}, we describe the process of achieving instant decal blending.

\subsection{Geometry Stage}\label{sec:geo_stage}

\begin{figure}[!t]
\centering
\includegraphics[width=0.98\columnwidth]{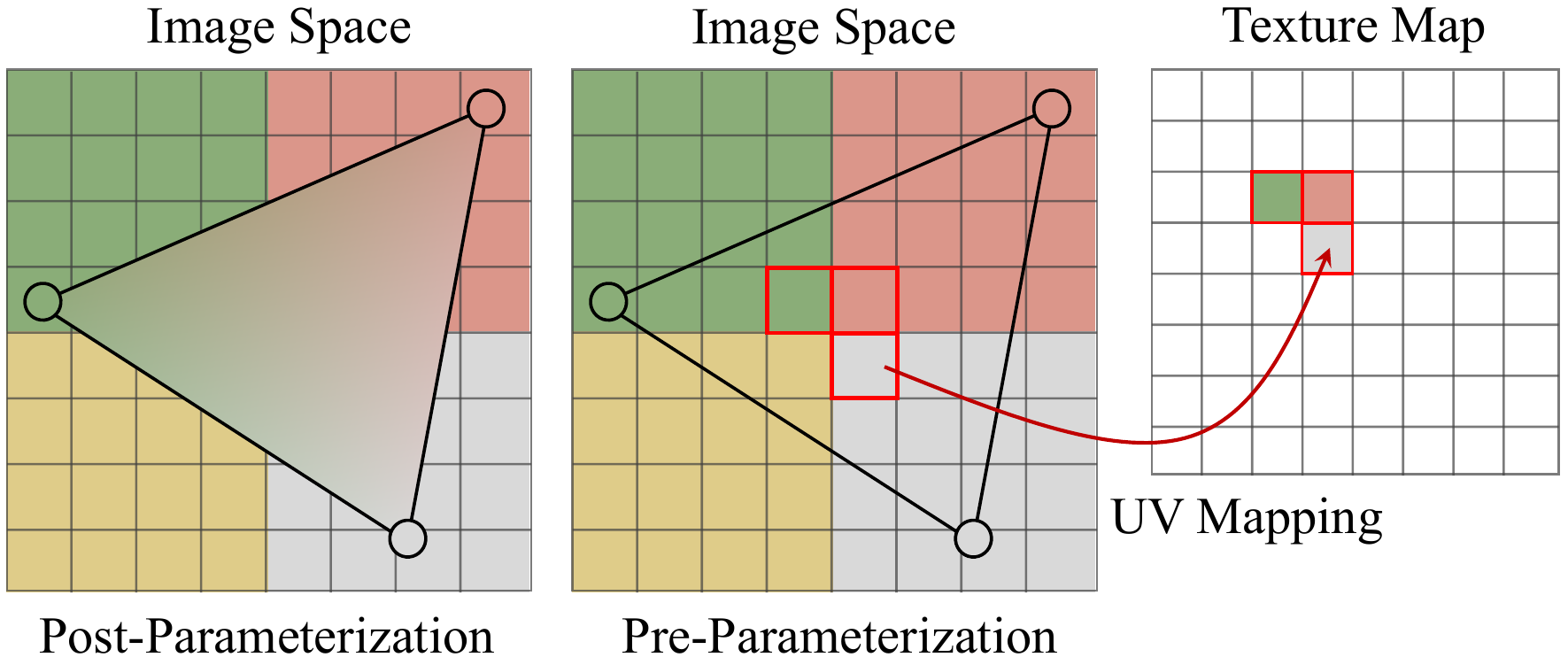}
\caption{
For post-parameterization, the rendered color within a triangle is limited to interpolating the features of its three vertices, hindering high-frequency texture representation. We address this by pre-parameterizing to obtain UVs and training a high-resolution texture map.
}
\label{fig_bary_motivation}
\end{figure}

We begin by utilizing advanced surface reconstruction techniques to generate high-quality meshes, specifically employing Voxurf \cite{wuvoxurf} for diffuse objects and Ref-NeuS \cite{ge2023ref} for specular objects. 
After generating the mesh, we select a continuous area, referred to as \textit{UV area}.
Each face of the mesh is assigned an indicator denoting whether it is included in UV area.
We then apply ARAP parameterization \cite{liu2008local} to UV area to obtain the UV coordinates for each vertex and allocate a sufficiently high-resolution texture map to store vertex features.
For rendering, we utilize rasterization techniques, a surface-based approach commonly employed in real-time applications. 
The color of each pixel is determined by the first-hit fragment, which includes the $z$-buffer, face index buffer, and barycentric coordinates. 
Following \cite{huang2023boosting}, we use the $z$-buffer to estimate the intersection coordinate $\mathbf{x} = (x, y, z)$.
Additionally, the ray direction $\mathbf{d}$ (which equals $-\omega_0$) is also stored in the fragments.
% zhang2023frequency

Following Lambertian and microfacet models, we assign each vertex the following learnable appearance features: diffuse albedo $\hat{\mathbf{a}}$, roughness $\hat{\rho}$, and metalness $\hat{m}$. 
Here, $\hat{\mathbf{a}}$ is a 3-dimensional vector, while both $\hat{\rho}$ and $\hat{m}$ are 1-dimensional.
In addition to these learnable features, each vertex also includes two fixed attributes: normal $\hat{\mathbf{n}}$ and UV coordinate $\hat{\mathbf{u}}$ (set to (0,0) for non-UV areas).
The fragment features are then calculated by performing barycentric interpolation, which yields a weighted sum of the vertex features.
To distinguish these interpolated fragment features from the original vertex features, we use the dot symbol to denote them. 
As illustrated in Fig. \ref{fig:model}, the final color of each pixel is computed using the interpolated fragment features: $\dot{\mathbf{a}}$, $\dot{\mathbf{u}}$, $\dot{m}$, etc.

\paragraph{Discussion of Rasterization.}
The following three reasons support our decision to use surface rasterization rather than volumetric representation employed in previous methods.
Firstly, existing surface reconstruction methods \cite{wuvoxurf, ge2023ref, fu2022geo} are capable of producing high-quality meshes, even for specular objects. 
This ensures that the contributions of samples beyond ray-surface intersections are negligible. Additionally, surface representations are more closely aligned with real-world objects than volumetric ones, which is crucial for achieving realistic sticker effects. 
Furthermore, since ray-surface intersections (or first-hit fragments) are guaranteed to lie on mesh triangles, we can utilize barycentric interpolation on vertex features, thereby avoiding the significant computational overhead of $k$-nearest neighbor searches.

\paragraph{Discussion of Pre-Parameterization.}
We plan to store the vertex diffuse albedo in a texture map using the UV mapping derived from ARAP, enabling surface-aligned texture editing.
As noted by \cite{do2016differential}, only surfaces that are simply connected and homeomorphic to a disk can be continuously unfolded.
Since most objects do not meet these criteria, they require cutting for unfolding \cite{srinivasan2023nuvo, xatlas}, which leads to fragmented texture spaces unsuitable for decal blending.
Therefore, we focus on parameterizing only a specific part of the mesh.
We refer to a naive method as \textbf{post-parameterization}, where the target area is parameterized after appearance disentangling is completed.
Specifically, this method converts the vertex diffuse albedo of the target area to a texture map using UV mapping, overlays the decal image onto this texture map, and then re-applies the edited texture back to the original vertices for rendering.
However, this approach may result in low-frequency details, as illustrated in Fig. \ref{fig_bary_motivation}. 
This occurs because a triangle can cover multiple pixels in image space, with the color determined by interpolating the three vertex colors, thereby limiting high-frequency texture detail.
To address this, we propose \textbf{pre-parameterizing} the UV area before appearance disentangling. For the UV area, each fragment diffuse albedo is indexed into the texture map using its interpolated fragment UV coordinates, allowing a single triangle to correspond to multiple pixels on the high-resolution texture map. 
During appearance disentangling, these albedos stored in the texture map can be correctly optimized, enabling a single triangle to express multiple colors, thus significantly enhancing the frequency representation capability.
Additionally, pre-parameterization enables instant editing without ARAP computation delays.

\subsection{Appearance Disentangling}\label{sec:AppearanceDisentangling}
\paragraph{Neural Albedo Texture Map.}
For UV area, we use a learnable texture map combined with fragment UV coordinates to represent the diffuse albedo, denoted as $\ddot{\mathbf{a}}$.
Regarding the learnable texture map, the naive approach is to directly optimize its pixels. 
However, this approach often results in incomplete learning of the texture map, as certain UVs may not be sampled during training.
Therefore, we propose a neural texture map implemented as an MLP that generates color values based on the input UV coordinates.
\begin{equation}
 \ddot{\mathbf{a}} = {F}_{t}(\gamma(\dot{\mathbf{u}})),
\end{equation}
where ${F}_{t}$ is a 3-layers MLP, $\gamma$ is a positional encoding function. 
% During the training process, for each iteration, the texture map is calculated by inputting a UV grid of shape $(H, W, 2)$, and will be precomputed during inference.
During training, at each iteration, the texture map is generated by inputting a UV grid of shape $(H, W, 2)$. This map will be precomputed for use during inference.

\paragraph{Diffuse Color Revising.}
In preliminary, we cover the pre-computation of specular color. 
Now, we will focus on the treatment of diffuse color in Eq. \ref{eq:diff_color}.
We point out that due to their differing physical properties, disentangling lighting and albedo from the diffuse color is significantly more challenging than disentangling specular component.
The view-dependent nature of specular reflection allows for a more robust decoupling of environmental lighting. 
In contrast, diffuse reflection, being view-independent, arises from multiple scattering and reflection of light within the surface, making it more challenging to separate environment lighting from the albedo.
Introducing environment lighting into the diffuse color can result in gradient descent baking the object’s albedo into the lighting, leading to erroneous optimization. 
% If environment lighting is introduced into the diffuse color, gradient descent may bake the object’s albedo into the environment lighting, leading to erroneous optimization results.
In many PBR techniques, the diffuse component typically excludes direct lighting and is combined with ambient lighting in a separate step, common in modern game engines and renderers \cite{pharr2016physically}.
Therefore, in our approach, we simplify the diffuse term Eq. \ref{eq:diff_color} by omitting the integral over the light source
\begin{equation}
\mathbf{c}_{\text {diffuse}}\approx\mathbf{a}(1-m).
\end{equation}

\begin{figure}[t]
\centering
\includegraphics[width=0.98\columnwidth]{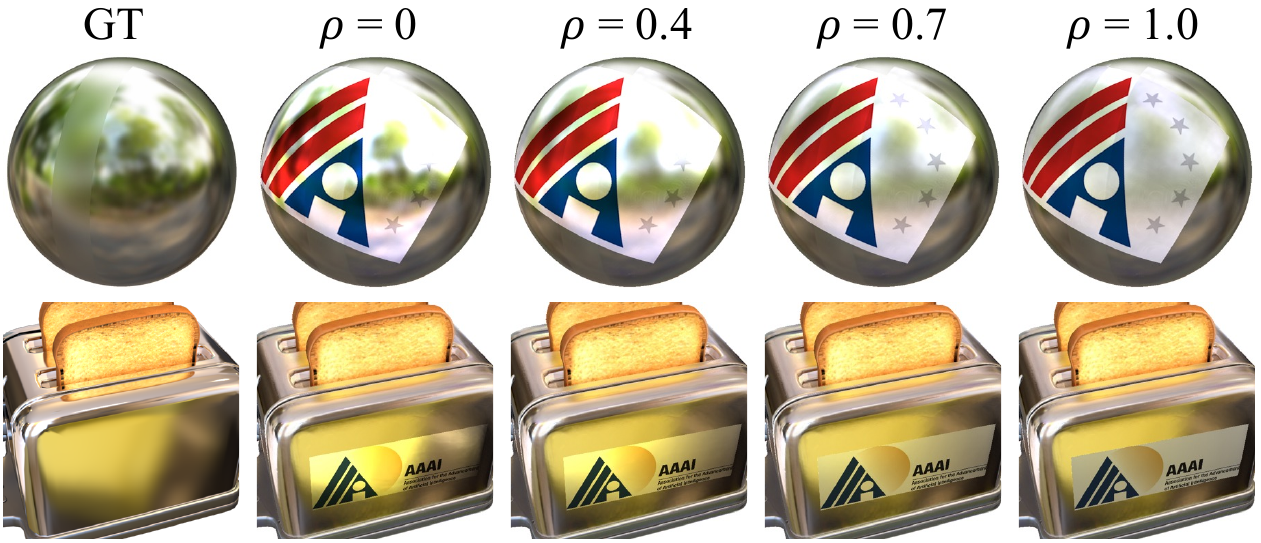}
% Reduce the figure size so that it is slightly narrower than the column. Don't use precise values for figure width. This setup will avoid overfull boxes.
\caption{
% Manually editing of roughness of the blending area, this can build diffuse effect on specular surface.
Manually editing the roughness of the blending area can create a diffuse effect on a specular surface.
% We can manually adjust the roughness of the blending area to blur the environment light, creating a diffuse sticker effect, 
}
\label{ex_rough_edit}
\end{figure}

\begin{figure*}[t]
\centering
\includegraphics[width=0.97\textwidth]{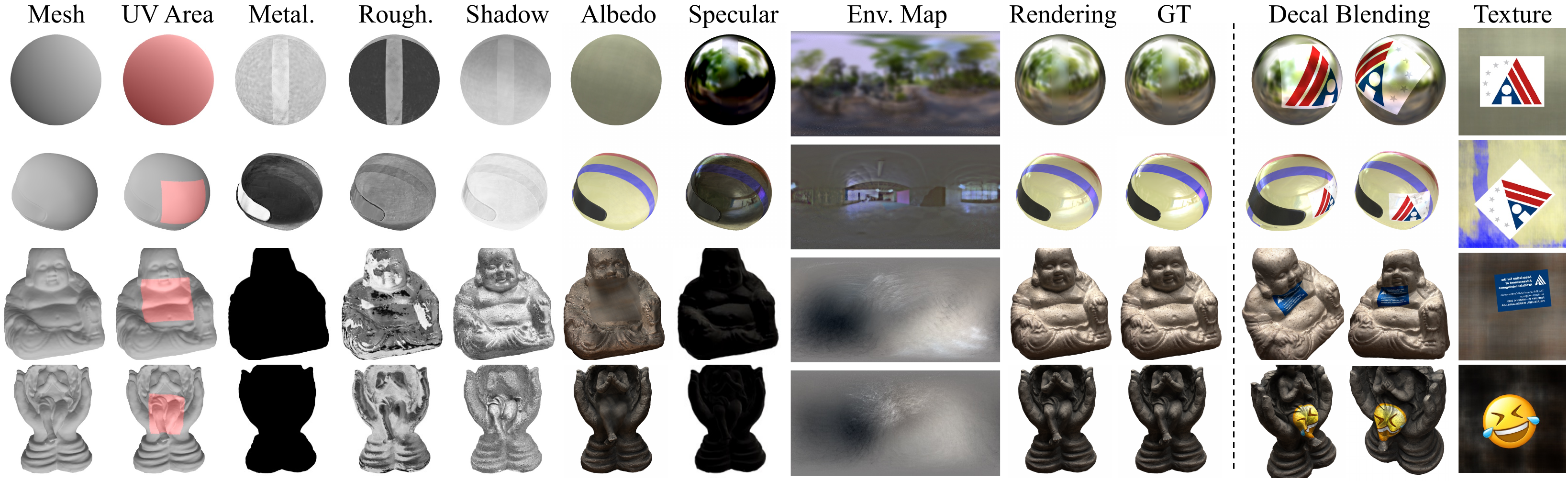}
% Reduce the figure size so that it is slightly narrower than the column. Don't use precise values for figure width. This setup will avoid overfull boxes.
\caption{
Disentangling and decal blending results of our method.
}
\label{fig_decouple_res}
\end{figure*}

\begin{figure*}[t]
\centering
\includegraphics[width=0.98\textwidth]{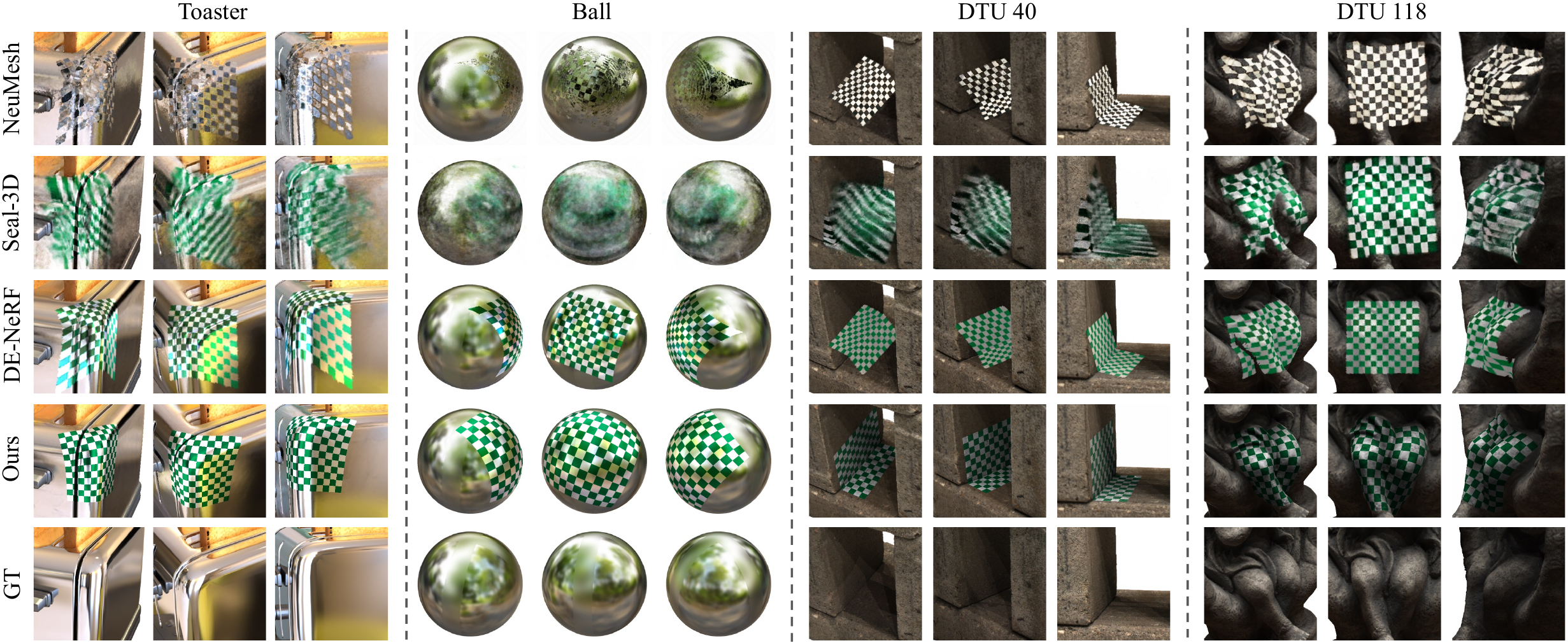}
% Reduce the figure size so that it is slightly narrower than the column. Don't use precise values for figure width. This setup will avoid overfull boxes.
\caption{
% We select a standard checkerboard pattern as the decal and apply it separately to two specular objects and two diffuse objects.
The results demonstrate that our method allows the decal to conform to the surface geometry with minimal warping, while the integration of lighting and shadows remains highly realistic.}
\label{ex_edit_compare}
\end{figure*}

\paragraph{Modeling Local Reflection.}
IBL employs a global environment map to simulate lighting, which generally yields favorable results.
However, objects with specular surfaces may reflect parts of themselves, making it challenging for the global environment map to accurately model these local reflections.
For example, the front panel of the \textit{Toaster} in Fig. \ref{fig:model} reflects the pattern of its base, a phenomenon we refer to as \textbf{local reflection}.
% and some metal material balls in \textit{Materials} reflect the patterns of other material balls; 
To address this, the following local reflection (LR) net is incorporated into the specular color (Eq. \ref{eq:spec_split}) to model the local reflection
\begin{equation}
\mathbf{c}_{\text {LR}}= F_r(\gamma(\mathbf{x}), \gamma(\mathbf{r})),
\end{equation}
in which $F_r$ is a 3-layers MLP, $\gamma$ is a positional encoding function, $\mathbf{x}$ is the intersection coordinate. 
Inspired by Ref-NeRF \cite{verbin2022ref}, we replace MLP's parameterization of view-dependent radiance with a representation of reflected radiance direction $\mathbf{r}=2\left(\omega_o \cdot \mathbf{n}\right)\mathbf{n}-\omega_o$,  
% $\omega_r=2\left(\omega_o \cdot n\right)-\omega_o$, 
instead of view direction.
In this way, Eq. \ref{eq:spec_split} is rewritten as
\begin{equation}\label{eq:spec_split_1}
% \mathbf{c}_{\text {specular}} = \mathbf{c}_{r}\cdot \underbrace{\int_{\Omega} L\left(\omega_i\right) D(\rho, \mathbf{r}) d \omega_i}_{L_{\text {specular}}} \cdot \underbrace{\int_{\Omega} \frac{D F G}{4\left(\omega_0 \cdot \mathbf{n}\right)} d \omega_i}_{M_{\text {specular}}}.
\mathbf{c}_{\text {specular}} = \mathbf{c}_{\text {LR}}\cdot L_{\text {specular}}\cdot M_{\text {specular}}.
\end{equation}

\paragraph{Modeling Shadow.}
Another limitation of IBL is its inability to model shadows, as it only considers direct lighting from the environment map.
This is often not an issue in reconstruction tasks, gradient optimization during reconstruction tends to bake shadows into the diffuse albedo.
However, this is particularly detrimental for decal blending, as edited textures lose the 3D perception provided by shadows, significantly reducing realism.
Inspired by ambient occlusion techniques \cite{zhukov1998ambient}, we aim to enable the model in multi-view reconstruction to adaptively learn an ambient occlusion factor that controls the brightness of each shading point, thereby creating a realistic 3D effect.
We refer to this factor as \textit{shadow factor}, denoted as $\tau$.
This 1-dimensional factor directly influences the final color, indicating the intensity of light received by the shading point.
We assign $\hat{\tau}$ to each vertex and use barycentric interpolation to obtain $\dot{\tau}$ for fragment.
Thus, Eq. \ref{eq:3} is rewritten as follows
\begin{equation}\label{eq:123}
\mathbf{c}\left(\omega_{\boldsymbol{o}}\right)= \sigma({\tau}) \cdot(\mathbf{c}_{\text {diffuse}}+\mathbf{c}_{\text {specular}}),
\end{equation}
where $\sigma$ is sigmoid function.
Since the albedo in the UV area is generated by an MLP, which is well-known for its strong spectral bias towards producing low frequency textures, whereas the shadow factor is not constrained by global frequencies and can capture higher frequency shadows.
In our experiments, we found that the shadow factor in the UV area can be adaptively optimized correctly without any supervision or constraints.

\paragraph{Pipeline.}
As shown in Fig. \ref{fig:model}, we optimize the parameters using only MSE loss, denoted as $L_2$, between the ground truth and the final color. 
The parameters being optimized include the vertex features, the parameters of ${F}_t$ and ${F}_r$, and an HDR environment map, represented as a float tensor.
% with the shape $(H{_\text{env}}, 2H{_\text{env}}, 3)$. 
Detailed information on the filtering process used to generate mipmaps from the environment map is provided in the appendix. 
During inference, we precompute ${F}_t$ to generate and cache the texture map for UV area, and similarly, the environment map is prefiltered and cached as a mipmap.

\subsection{Decal Blending} \label{sec:decal_blending}
Once the reconstruction is complete, we can upload an image and select four anchors in UV area to define a quadrilateral in the texture space. 
The uploaded image is then directly overlaid onto the corresponding position in the precomputed albedo texture map, instantly achieving the desired 3D editing effect. 
Please see the appendix for details. 
The decal appears like a sticker on diffuse surfaces, while on metal surfaces, it naturally blends with environment specular reflections. 
We can also manually adjust the roughness of the blending area to blur the environment light, creating a diffuse sticker effect, as shown in Fig. \ref{ex_rough_edit}.

\section{Experiments}
\subsection{Experimental Settings}
\paragraph{Implementation Details.}

For convenience, we use PyTorch3D \cite{ravi2020accelerating} to pre-rasterize the mesh and store the fragments locally, with OpenGL \cite{kessenich2016opengl} employed for deployment. 
The Adam optimizer is used, with a learning rate of 5e-4 for $F_t$ and $F_r$, and 5e-3 for the appearance features and environment map. 
Training is performed on a single NVIDIA RTX 3090 GPU and takes only 1 hour. 
The positional encoding dimension is set to 4, and the hidden layer dimension of MLP is 256. 
We utilize ARAP interface provided by Libigl library \cite{jacobson2017libigl}. 
% More details can be found in appendix.
\paragraph{Datasets and Baselines.}
We evaluate our method on NeRF-Synthetic \cite{mildenhall2020nerf}, Shiny Blender \cite{verbin2022ref}, and DTU \cite{jensen2014large} dataset.
We mainly compare our method with the following methods capable of performing decal blending on reconstructed objects: NeuMesh, Seal-3D, and DE-NeRF.

\subsection{Reconstruction Results}
We present the components of disentangling and editing results in Fig. \ref{fig_decouple_res}. 
Since disentangled reconstruction is more challenging than free-form reconstruction, we compare our approach with advanced disentangled reconstruction methods for fairness, in terms of PSNR.
The results demonstrate that our reconstruction quality is comparable to that of DE-NeRF, as shown in Tab. \ref{tab:rec_comparison}. 
Comparison results in terms of SSIM and LPIPS can be found in the appendix.

\subsection{Editing Results}
\paragraph{Qualitative Comparison.} 
For the qualitative evaluation of decal blending, we select a standard checkerboard pattern as the decal, apply it to two specular objects and two diffuse objects, and compare our method with three previous decal blending methods.
The decal in NeuMesh appears similar in color to the object because the decoder is fixed, and only the vertex features are optimized.
Seal3D, using a volumetric representation, fails to accurately locate the surface, making the decal appear embedded within the surface. 
Although DE-NeRF achieves decent lighting integration on specular objects, it fails to handle shadow effects. 
Moreover, all three methods use a manually edited single-view image as the optimization target, leading to significant warping.
In contrast, our method not only incorporates environment lighting and shadows on decals but also maintains geometric alignment.

\paragraph{Quantitative Comparison.}
% We design the Ratio Variance Warping (RVW) metric to assess decal area warping. 
The Ratio Variance Warping (RVW) metric is designed to assess decal area warping. 
We sample $N$ vertex pairs in the UV area, forming set $V$, and map them to 2D texture space as set $U$. 
Geodesic distances $g_i$ in $V$ and Euclidean distances $d_i$ in $U$ are used to compute ratios $r_i = \frac{g_i}{d_i}$. 
The variance $\text{Var}(r_i)$ measures warping, with zero variance indicating uniform scaling.
Additionally, to further evaluate the practicality of each method, we compare training time, rendering FPS, and editing time. 
As shown in Tab. \ref{tab_all}, our method not only minimizes decal warping but also significantly surpasses previous methods in terms of practicality.
Furthermore, we conduct a user study with 18 participants to evaluate the integration of decal texture with lighting and shadows. 
We ask the participants to give a preference score (ranging from 0 to 1.0), with 1.0 indicating that the integration appears completely natural and 0 indicating no environmental lighting effect.
Our score is \textbf{0.86}, while the second-place method, DE-NeRF, scores 0.55.

\begin{table}[t]
\centering
\begin{tabular}{lccc}
\toprule
Method & NeRF Syn.  & Shiny Blen.  & DTU\\
\midrule
PhySG & 20.60 & 26.21 & 24.69 \\
NeRFactor & 27.86 & 27.04 & 25.75 \\
NvDiffRec & 29.05 & 28.11 & 25.54 \\
DE-NeRF & \textbf{29.18} & 28.79 & 25.93 \\
Ours & 28.96 & \textbf{28.81} & \textbf{26.12} \\
\bottomrule
\end{tabular}
\caption{PSNR of novel view synthesis with disentangled reconstruction methods.}
\label{tab:rec_comparison}
\end{table}

\begin{table}[t]
\centering
    \begin{tabular}{lcccc}
        \toprule
        % \multicolumn{3}{c}{~} &\multicolumn{2}{|c|}{\small{Editing Ability}} & \multicolumn{3}{|c|}{NeRF-Synthetic}& \multicolumn{3}{|c}{Tanks\&Temples}\\ 
        % \hline
        Method & Training$\downarrow$  & FPS$\uparrow$ & Editing$\downarrow$ & Warping$\downarrow$ \\
        \hline
        NeuMesh & 16 hours & 0.017 & 1 hour & 0.472 \\
        Seal-3D & \textbf{0.1 hours} & 28.4 & 1 min & 0.565 \\
        DE-NeRF & 18 hours & 0.013 & 1 hour & 0.489 \\
        Ours & 1 hour & \textbf{81.5} & \textbf{0} & \textbf{0.001} \\
        \hline
    \end{tabular}
    \caption{
    We compare training time, inference frame rate (FPS), editing time, and the RVW metric for warping. 
    While our training time is longer than Seal-3D, our method significantly outperforms other methods  in the other three metrics.
    }
    \label{tab_all}
\end{table}

\begin{figure}[!t]
\centering
\includegraphics[width=0.98\columnwidth]{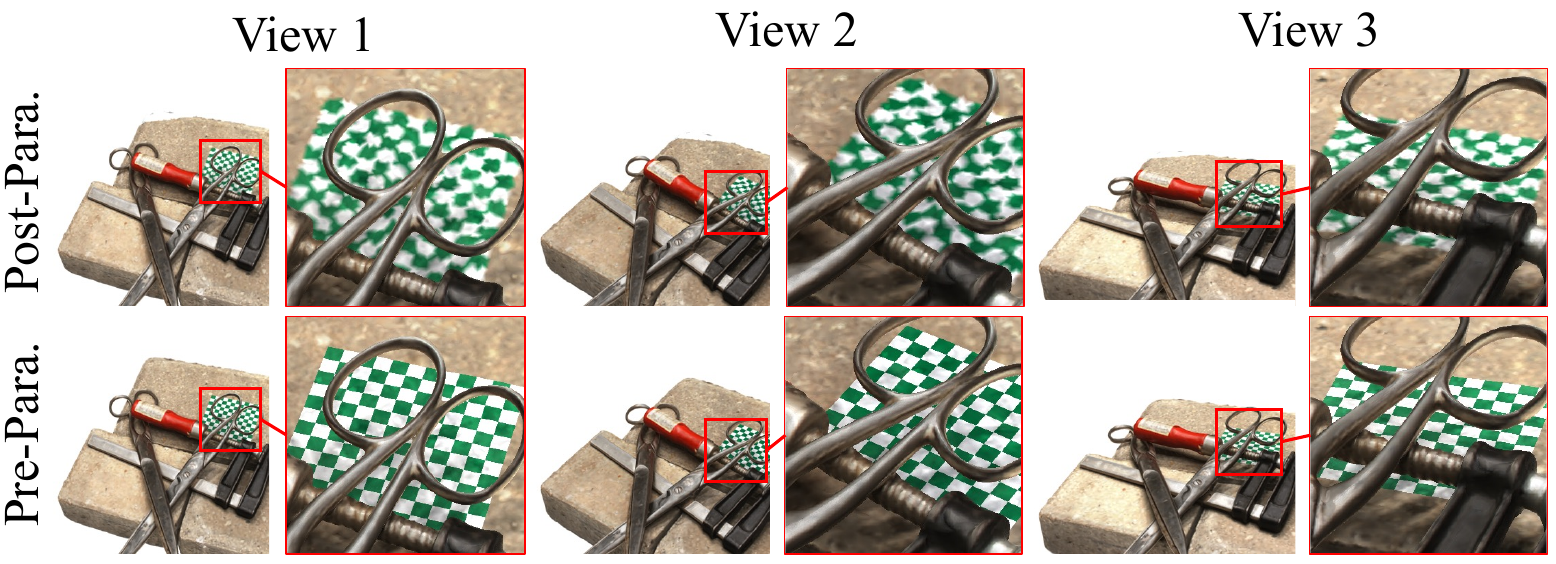}
% Reduce the figure size so that it is slightly narrower than the column. Don't use precise values for figure width. This setup will avoid overfull boxes.
\caption{Our proposed pre-parameterizaton shows high-frequency details of the decal (i.e., the cell edges are clear), while the other one exhibits significant blurriness.}
\label{ex_pre_vs_post}
\end{figure}

\begin{figure}[t]
\centering
\includegraphics[width=0.98\columnwidth]{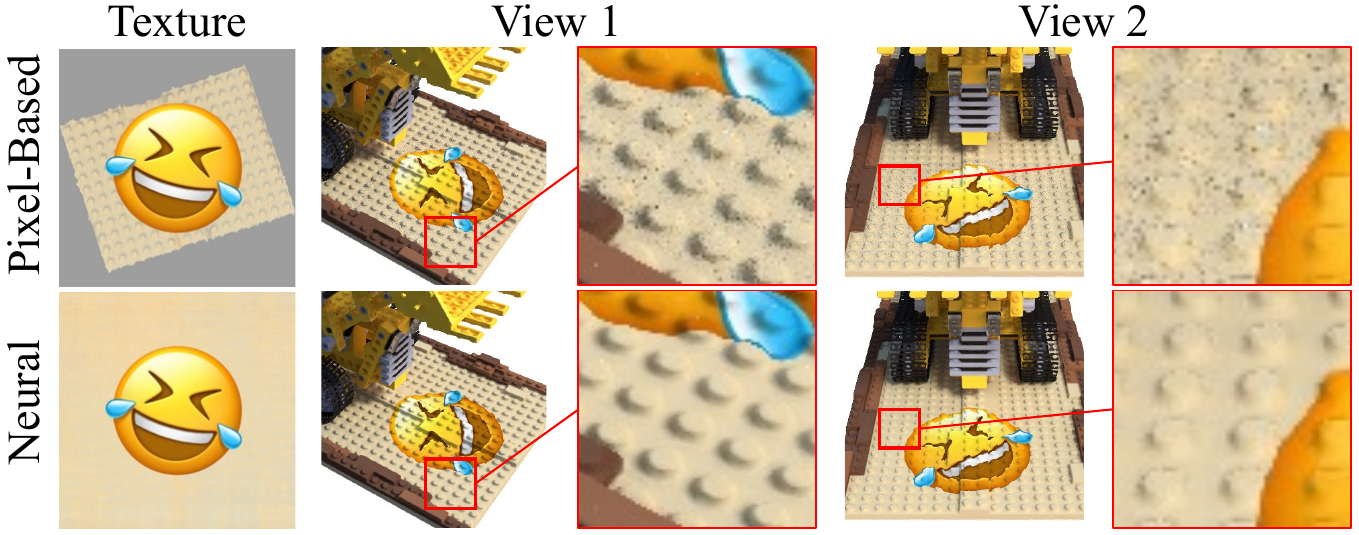}
% Reduce the figure size so that it is slightly narrower than the column. Don't use precise values for figure width. This setup will avoid overfull boxes.
\caption{Pixel-based texture maps result in insufficient sampling, leading to noisy renderings in novel views.}
\label{ex_abla_lego_mlp}
\end{figure}

\begin{figure}[!t]
\centering
\includegraphics[width=0.98\columnwidth]{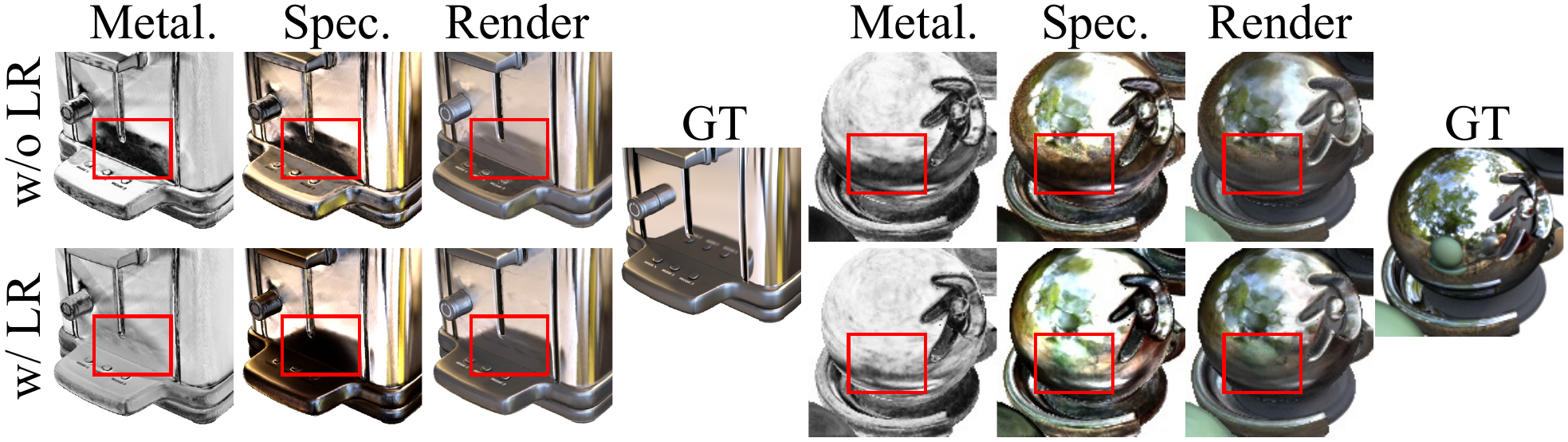}
% Reduce the figure size so that it is slightly narrower than the column. Don't use precise values for figure width. This setup will avoid overfull boxes.
\caption{LR net can model local reflections within objects, improving the quality of the reconstruction.}
\label{ex_abla_IR_mat_toas}
\end{figure}

\begin{figure}[!t]
\centering
\includegraphics[width=0.98\columnwidth]{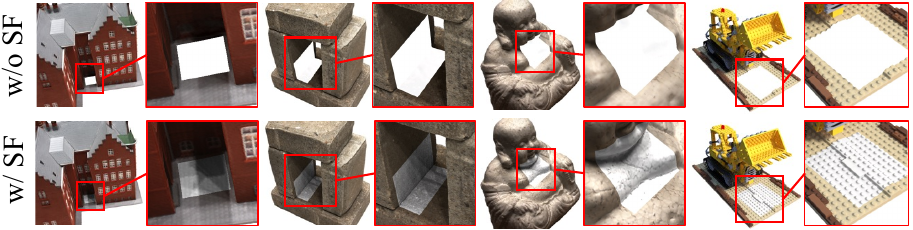}
% Reduce the figure size so that it is slightly narrower than the column. Don't use precise values for figure width. This setup will avoid overfull boxes.
\caption{
Our proposed shadow factor (SF) accurately captures shadow brightness, enhancing the white decal’s three-dimensional appearance and sense of depth.
}
\label{ex_abla_shadow_f}
\end{figure}

\subsection{Ablation Study}
In this section, we evaluate several technical components in our pipeline through a series of ablation studies.
\paragraph{Pre-Parameterization.}
As shown in Fig. \ref{ex_pre_vs_post}, the baseline model with post-parameterization (as discussed in Sec. \ref{sec:geo_stage}) exhibits significant blurriness on the decal. 
In contrast, our method shows high-frequency details of the decal (i.e., the cell edges are clear), as previously analyzed in Fig. \ref{fig_bary_motivation}.
\paragraph{Neural Albedo Texture Map.}
We use the optimization of the pixels' RGB of texture map as a baseline (pixel-based), comparing it to our approach, which encodes textures using an MLP.
As shown in Fig. \ref{ex_abla_lego_mlp}, the baseline approach, which lacks sufficient UV sampling during training, leads to noise artifacts in novel views. 
By contrast, encoding the texture map using an MLP allows the model to generalize across the entire UV space from supervision on a subset of pixels.

\paragraph{LR Net.}
We remove LR net from our pipeline to create a baseline.
As shown in Fig. \ref{ex_abla_IR_mat_toas}, our method reconstructs the local reflections, whereas the baseline method fails.
\paragraph{Shadow Factor.}
We remove shadow factor from Eq. \ref{eq:123} to build a baseline and use a plain white image as the decal to highlight the shadow effect. 
As shown in Fig. \ref{ex_abla_shadow_f}, our proposed method effectively decouples shadows and projects them onto the decal, achieving stereoscopic impression.
\section{Conclusion}
In this paper, we present InstantSticker, a disentangled reconstruction pipeline based on IBL, which focuses on highly realistic decal blending, simulates stickers attached to the reconstructed surface, and allows for instant editing and real-time rendering.
Extensive experimental results demonstrate that our method surpasses previous decal blending methods in terms of editing speed, rendering speed, and editing quality, achieving the state-of-the-art.
\paragraph{Limitation.}
Our method is dependent on high-quality mesh input, a limitation we plan to address in future work by exploring approaches that relax this requirement.

\section*{Acknowledgments}
This work was supported by National Science Foundation of China (U20B2072, 61976137). This work was also partially supported by Grant YG2021ZD18 from Shanghai Jiao Tong University Medical Engineering Cross Research. 
This work was partially supported by STCSM 22DZ2229005.

\bibliography{aaai25}

\end{document}